\documentclass[10pt,twocolumn,letterpaper]{article}

\usepackage{iccv}
\usepackage{times}
\usepackage{subfigure}
\usepackage{epsfig}
\usepackage{graphicx}
\usepackage{amsmath}
\usepackage{amssymb}

\usepackage[breaklinks=true,bookmarks=false]{hyperref}

\iccvfinalcopy 

\def\iccvPaperID{17} 

\begin{document}

\title{Small Obstacle Avoidance Based on RGB-D Semantic Segmentation}


\author{Minjie Hua$^*$\\
CloudMinds Technologies\\
{\tt\small michael.hua@cloudminds.com}
\and
Yibing Nan\thanks{\ indicates equal contribution}\\
CloudMinds Technologies\\
{\tt\small charlie.nan@cloudminds.com}
\and
Shiguo Lian\\
CloudMinds Technologies\\
{\tt\small sg\_lian@163.com}
}

\maketitle

\begin{abstract}
  This paper presents a novel obstacle avoidance system for road robots equipped with RGB-D sensor that captures scenes of its way forward. The purpose of the system is to have road robots move around autonomously and constantly without any collision even with small obstacles, which are often missed by existing solutions. For each input RGB-D image, the system uses a new two-stage semantic segmentation network followed by the morphological processing to generate the accurate semantic map containing road and obstacles. Based on the map, the local path planning is applied to avoid possible collision. Additionally, optical flow supervision and motion blurring augmented training scheme is applied to improve temporal consistency between adjacent frames and overcome the disturbance caused by camera shake. Various experiments are conducted to show that the proposed architecture obtains high performance both in indoor and outdoor scenarios.
\end{abstract}

\section{Introduction}

\begin{figure} [tbp]
  \centering
  \subfigure[]{
    \centering
    \label{fig:case:a} 
    \begin{minipage}[b]{0.15\textwidth}
      \centering
      \includegraphics[width=1\linewidth]{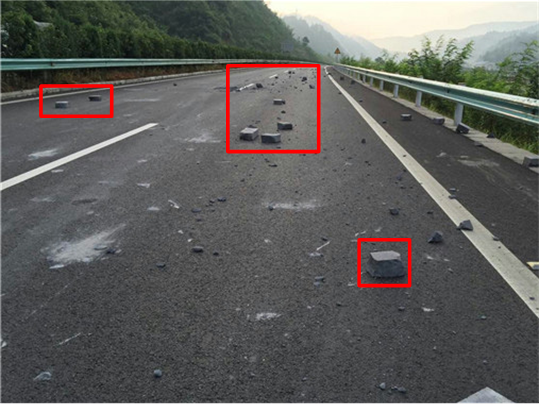}
    \end{minipage}}
  \subfigure[]{
    \centering
    \label{fig:case:b}
    \begin{minipage}[b]{0.15\textwidth}
      \centering
      \includegraphics[width=1\linewidth]{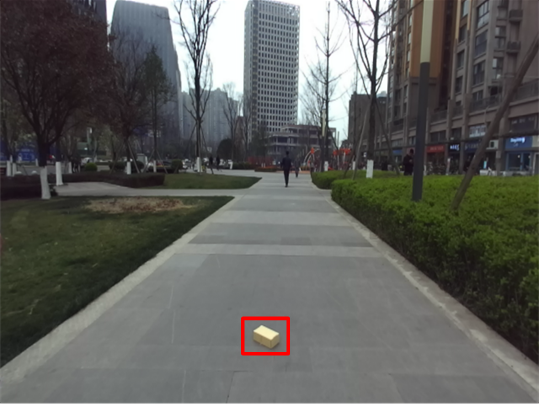}
    \end{minipage}}
  \subfigure[]{
    \centering
    \label{fig:case:c} 
    \begin{minipage}[b]{0.15\textwidth}
      \centering
      \includegraphics[width=1\linewidth]{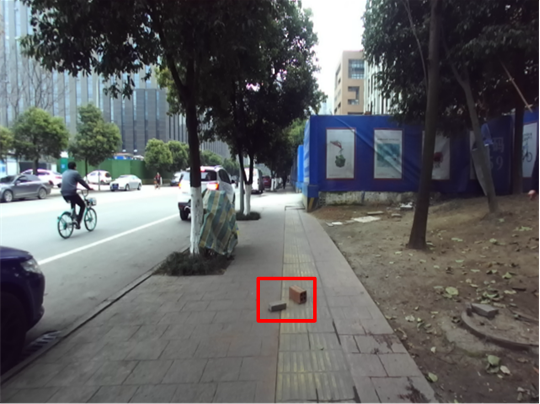}
    \end{minipage}}
  \caption{Some cases with small obstacles in autonomous driving (a), patrol robot (b) and blind guidance (c).}
  \label{fig:case}
\end{figure}

  Obstacle avoidance is a fundamental component for intelligent mobile robots, the core of which is to have an environmental perception module that helps identify the possible hindrance that can block the way of the robots. Especially, for some scenarios, small obstacles should be carefully considered, \eg, autonomous driving, patrol robot and blind guidance, as shown in Fig.~\ref{fig:case}. In autonomous driving, while the car running at a high speed, such small obstacle as a brick on road may cause the car to turn over. In blind guidance, visually impaired people are fragile to such small object even only 3cm higher on road. For the patrol robot working as a community policing, some remnants scattered on road should be detected either as obstacles to avoid or spilled garbage to alert, such as garbage cans and stone blocks.

  Range-based and appearance-based methods are two major approaches to perform the task of obstacle detection. But the former share disadvantages, where it is difficult to detect small obstacles and distinguish between diverse types of road surfaces, making it hard to identify the sidewalk pavement from adjacent grassy area, a combination commonly seen under urban circumstances.

  The appearance-based methods~\cite{NJ84,HI95,TM87,LB97,UN00}, on the other hand, are not subjected to the above problems. Since they define obstacles as objects that differ in appearance, which is more essential a character in our study, from the road, while the former only define obstacles as objects that rise at a certain height from the road. One of the main procedures of appearance-based obstacle avoidance methods can be identified as semantic segmentation of accessible area and other objects such that the robots can make decisions of obstacle avoidance and path planning.

  For appearance-based sensors, we have both the monocular camera and the stereo camera. The former outputs pure visual cues (\ie RGB information), where it is difficult for semantic segmentation networks to accurate distinguish real obstacles and appearance changes such as different road colors and road markings. The latter provides additional depth channel to RGB images. Incorporating RGB and depth information could potentially improve the performance of semantic segmentation networks, which is suggested by recent efforts~\cite{JZ17,HM16,JZ18}.

  Utilizing semantic segmentation to obstacle avoidance is a new topic in recent years~\cite{KK16,RG17,GJ18}, but these methods are applied to limited scene or not sensitive to small obstacles. In this paper, we propose a novel small obstacle avoidance system based on RGB-D semantic segmentation that can be widely used in indoor and outdoor scenarios.

  The main contributions of this work are as follows: 1). An two-stage RGB-D encoder-decoder network is proposed for obstacle segmentation, which segments the image to get the road mask first and then gets more accurate obstacle region even small obstacle region from the extracted road area. 2) An optical flow supervision between adjacent frames is proposed for the segmentation network to keep the temporal consistency, which is critical for stable obstacle detection when camera moves. 3) A motion blur based data augmentation scheme is proposed to suppress segmentation errors introduced by camera shake. 4) A small obstacle avoidance system based on RGB-D semantic segmentation is proposed and evaluated based on collected practical datasets both of indoor and outdoor scenarios.

\section{Related Work}

\subsection{Obstacle Detection and Avoidance}

 The first autonomous mobile robot used the appearance-based method for obstacle detection~\cite{NJ84}, named Shakey, developed by Nilsson. It can detect obstacles by edge detection on the monochrome image. Following Nilsson's step, Horswill also applied edge detection method to his mobile robot named Polly~\cite{HI95}, which was operated in real-life environment. The edge detection method that Shakey and Polly used only performed well when the floor had little texture and the environment was uniformly illuminated. It's very sensitive to floor color variation and lighting conditions.

 Besides the edge information, some other information have also been used for obstacle detection. Turk and Marra made use of the color information~\cite{TM87} to detect obstacles by subtracting consecutive frames of a video. Lorigo proposed an algorithm using both color information and edge information that worked on texture floor~\cite{LB97}. Lorigo assumed there is no obstacle right in front of the robot and used this area as reference area to find obstacle. Ulrich improved Lorigo's algorithm with a candidate queue and a reference queue~\cite{UN00}. However, before the mobile robot can move autonomously, they need to steer it first for several meters to form a correct reference area. And if the road outside the reference area differs from the reference area due to unexpected shadow or reflections, it will be incorrectly classified as an obstacle as well.

 Recently, some other appearance based obstacle avoidance schemes have been proposed. Ghosh and Wei proposed stereo vision based methods to detect obstacles ~\cite{GB17,WG14}. In these methods, the disparity map was used to analyze surroundings and determine obstacles. However, the disparity map computed by stereo images is not robust enough in complex environments especially in the presence of small obstacles. Yang proposed a blind guiding system based on both RGB and depth images~\cite{YK18}. In this method, the RGB image was used to get semantic map of the environment, and then combined with depth image to percept the terrain. However, this method focuses on the segmentation of transitable areas including pedestrians and vehicles, while it is incapable of detecting small obstacles. For small obstacles, \cite{RG17,GJ18} explored the obstacle detection issue focusing on autonomous driving scenarios. In these methods, pre-defined obstacle categories that are common in driving scenarios were placed on road to initialize the obstacle dataset first,and then RGB and depth information were both used to train the obstacle segmentation models. However, the method supports only some normal obstacles predefined on traffic road while not extensible for arbitrary obstacles on road.

\subsection{Semantic Segmentation}

 Recently, there have been some advances on deep neural networks based semantic segmentation. The task of semantic segmentation is to label each pixel of an image with a semantic class. FCN~\cite{FCN}, the pioneer work in the exploration of CNN-based semantic segmentation methods, adapts classifiers for dense prediction by replacing the last fully-connected layer with convolution layers. SegNet~\cite{SEGNET} is an classical encoder-decoder architecture and reuses the pooling indices from the encoder to decrease parameters. DeepLab~\cite{DLAB1,DLAB2,DLAB3} proposes atrous spatial pyramid pooling (ASPP) for exploiting multi-scale information and then augments the ASPP module with image-level feature to capture longer range information. PSPNet~\cite{PSPNET} performs spatial pooling at several grid scales and demonstrates excellent performance. ICNet~\cite{ICNET} achieves great balance between efficiency and accuracy by using a hierarchical structure to save time on high-resolution feature maps.

\begin{figure}[bp]%
\centering
\includegraphics[width=0.85\linewidth]{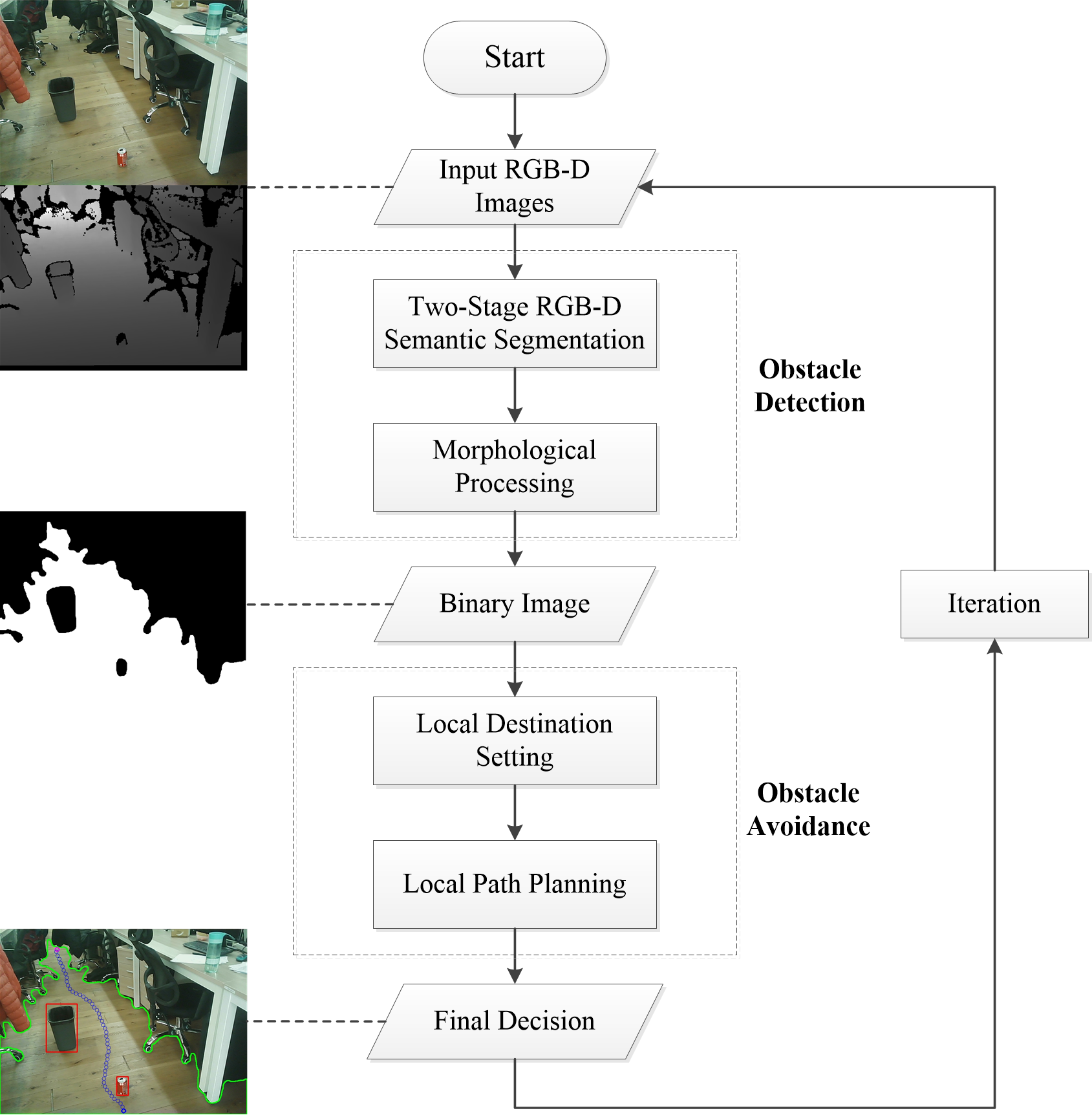}
\caption{The flow chart of proposed system.}
\label{fig:flow}
\end{figure}

\begin{figure*}[tbp]%
\centering
\includegraphics[width=0.83\linewidth]{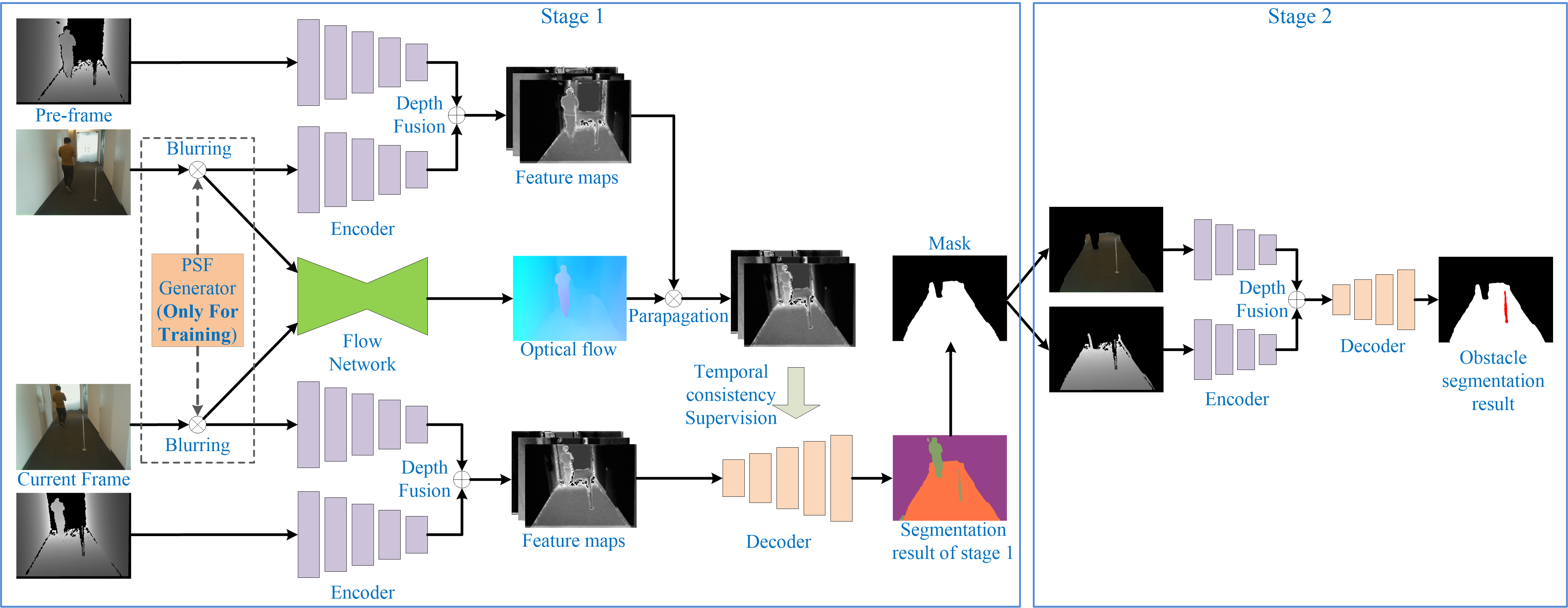}
\caption{The proposed two-stage RGB-D semantic segmentation network architecture.}
\label{fig:archi}
\end{figure*}

 As regards RGB-D semantic segmentation, some studies have tried to utilizing the depth information to achieve better segmentation accuracy~\cite{HM16,JZ18,MS17,VM18}. Hazirbas~\cite{HM16} presented a fusion-based CNN architecture which is consisted of two encoder branches for RGB and depth channel. The features of two branches are fused on different layers. Based on the fusion-based encoder-decoder architecture, Jiang~\cite{JZ18} applied pyramid supervision training scheme and got a good result.

 Most of existing semantic segmentation schemes do not consider the condition of practical applications. For example, the camera often shakes during robot moving and thus captures the image sequence with shaking and/or blurs, which may lead to wrong segmentation results. And in actual application, pixel values and segmentation results of adjacent frames often vary greatly even though the camera and target don't move at all. This is determined by the hardware characteristics of imaging sensor and nonlinearity of deep neural networks. What's more, the original image contains abundant semantic information, making it hard to give specific definition of road and obstacle directly for one-stage segmentation models. In this paper, we adapt the RGB-D fusion-based structure, and introduce optical flow supervision and motion blurring training scheme to improve temporal consistency between adjacent frames. Moreover, a two-stage semantic segmentation architecture is proposed to get accurate semantic result for road and obstacle.

\section{Approach}

 As illustrated in Fig.~\ref{fig:flow}, the proposed obstacle detection and avoidance system contains several steps: RGB-D based two-stage semantic segmentation, morphological processing, local destination setting and path planning. The two-stage semantic segmentation transforms the input RGB-D image to a raw binary image, which is then smoothed in morphological processing. As a result, the module generates a calibrated binary image where every pixel is labelled as either road or obstacle. Then the binary image is passed to the obstacle avoidance module to determine a destination and a walkable path. The whole process works repeatedly during robot's moving.

\subsection{Two-Stage RGB-D Semantic Segmentation}

 The architecture of proposed two-stage RGB-D semantic segmentation is shown in Fig.~\ref{fig:archi}. For the first stage, the RGB-D encoder part of the network is similar with previous work of RedNet~\cite{JZ18}, which has two convolutional branches. Both of the RGB and depth branches adopt ResNet architecture that with global average pooling layer and fully-convolutional layer removed. Different scale of features are extracted for RGB and depth channel respectively. For each convolutional operation, batch normalization is performed before ReLU function. Feature maps from two branches are fused at each scale. The fusion operation is denoted as ${{f}_{RGB}}\oplus {{f}_{D}}$, where ${{f}_{RGB}}$ denotes feature maps of the RGB branch and ${{f}_{D}}$ denotes feature maps of the depth branch, $\oplus$ denotes direct element-wise summation. Two consecutive video frames ${{I}_{\text{p}}}$ and ${{I}_{\text{c}}}$ are input to the encoder at the same time, where ${{I}_{\text{p}}}$ and ${{I}_{\text{c}}}$ denotes the pre-frame and the current frame. The RGB channel of  ${{I}_{\text{p}}}$ and ${{I}_{\text{c}}}$ are fed to a deep flow network to estimate their optical flow field ${{F}_{c\to p}}$ in the first stage.

 Semantic information and their spatial location relationship is encoded in the feature maps. Consecutive video frames have highly similarity, so we can propagate feature maps of the pre-frame back to current frame through their flow field~\cite{ZX17}. And the propagated feature maps should be also highly similar with feature maps generated by original current frame. In this work, we choose the FlowNet~\cite{FLOWNET} inception architecture which is modified in~\cite{ZX17} to meet the tradeoff between accuracy and speed. Because the feature maps have different spatial resolution, the flow field is bi-linearly resized to the same resolution of the feature maps for propagation. So, given a position $x$ on a pre-frame feature map ${{f}_{p}}$, the propagated value can be represented as:
\begin{equation}\label{eq:fcx}
\begin{aligned}
  {{f}_{c}}(x)={{S}_{c\to p}}(x)\sum\limits_{i}{B(i,x+{{F}_{c\to p}}(x)){{f}_{p}}(i)}
\end{aligned}
\end{equation}
 where $B$ denotes the bilinear interpolation kernel, ${{S}_{c\to p}}$ denotes the pixel-wise scale function introduced in~\cite{ZX17}.

\begin{figure}[tbp]%
\centering
\includegraphics[width=0.65\linewidth]{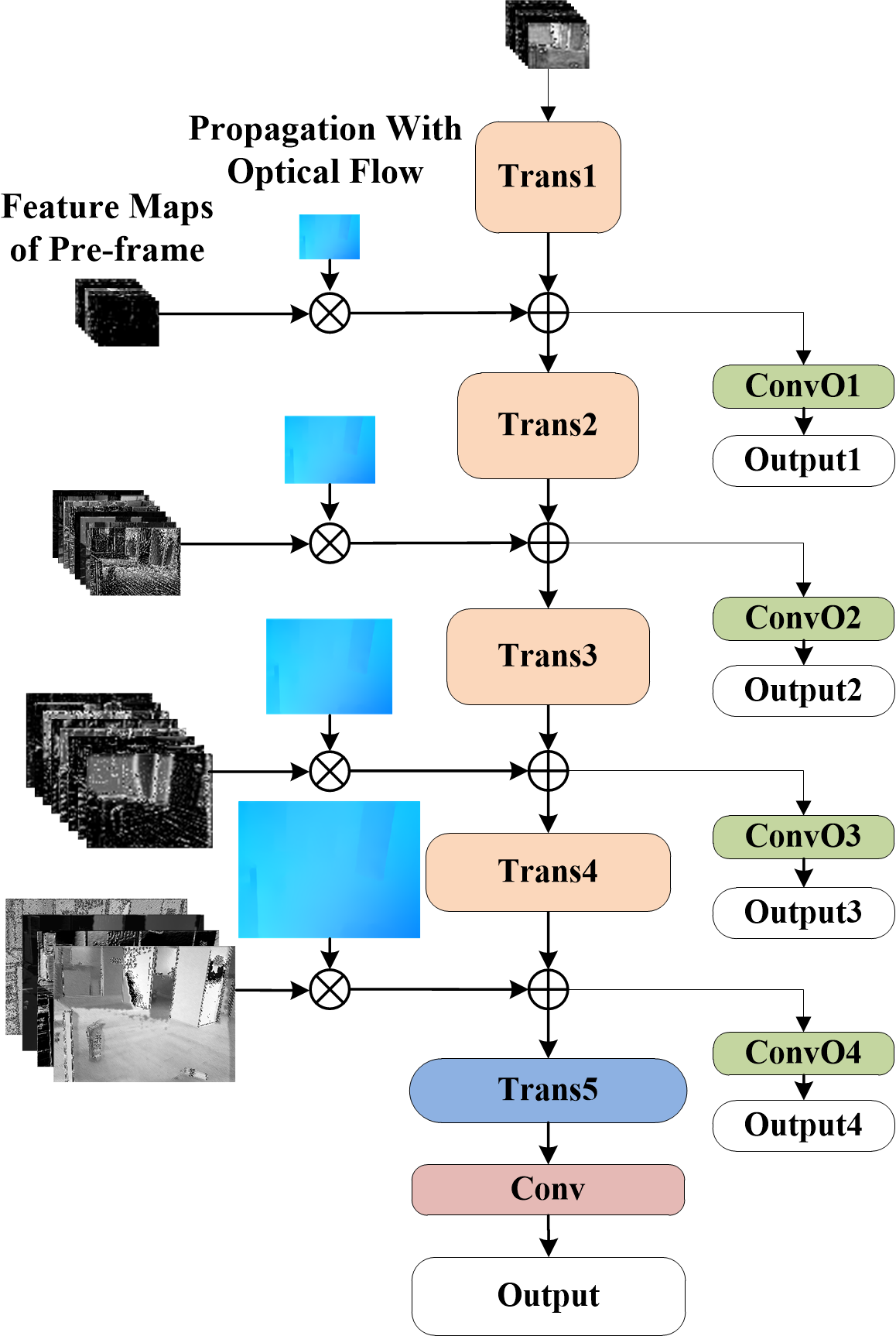}
\caption{Decoder architecture of the first stage.}
\label{fig:decod}
\end{figure}

 For the decoder, different with the original RedNet, the inputs of skip connection are replaced by the propagated feature maps to keep temporal consistency of segmentation results between adjacent frames. As illustrated in Fig.~\ref{fig:decod}, each scale of feature map, except the last layer of encoder of the ${{I}_{\text{p}}}$ branch, is propagated to current frame and fused to the transpose convolution layer of ${{I}_{\text{c}}}$. The fused method is also element-wise summation. Then the fused feature maps are fed into convolution layers with $1\times 1$ kernel size and stride one to generate score maps, \ie output1, output2, output3, and output4. These four side outputs together with the final output are fed into a softmax layer to get their classification score of semantic classes. Loss function for each output is formulated by calculating cross entropy loss of the classification score map.
\begin{equation}\label{eq:loss}
\begin{aligned}
  Loss({{W}_{1}})=-\frac{1}{N}\sum\limits_{i}{\log \left( \frac{\exp ({{s}_{{{g}_{i}}}}(i;{{W}_{1}}))}{\sum\nolimits_{k}{\exp ({{s}_{k}}(i;{{W}_{1}}))}} \right)}
\end{aligned}
\end{equation}
 where ${{W}_{1}}$ denotes the network parameters of stage one, $i$ is the pixel location, and $N$ denotes the spatial resolution of corresponding output, $s$ denotes the score map, and $g_i$ is the ground truth of class number on location $i$. The overall loss of the stage one network is calculated by adding the five losses together. For the ground truth has full resolution, it's downsampled to adapt the resolution of side outputs.

 After semantic segmentation of stage one, we can extract outer contour of road area by pixel labels. Then the contour \ie ROI is mapped to the original RGB-D image. Pixels out of the contour are set to zeros to eliminate the influence of these object semantic information out of the contour during training phase. It should be noted that non-road pixels within the contour are remain unchanged. The processed ROI of RGB-D image is then fed to the second stage of semantic segmentation. We define this ROI as:
\begin{equation}\label{eq:mask}
\begin{aligned}
Mas{k_{seg \to RGB - D}}(i) = \left\{ \begin{array}{l}
1,\quad \text{if}{\;}i{\rm{ }} \in {\rm{ ROI}}\\
0,\quad \text{otherwise}
\end{array} \right.
\end{aligned}
\end{equation}

 In the second stage, there are three categories of semantic concept: road, obstacle on the road and others. Objects within road contours are all mapped to obstacle class during training. The last output is:
\begin{equation}\label{eq:obs}
\begin{aligned}
  Obstacl{{e}_{seg}}={{W}_{2}}({{I}_{c}}\otimes Mas{{k}_{seg\to RGB-D}})
\end{aligned}
\end{equation}
 where ${{W}_{2}}$ denotes network parameters of stage two, $Obstacl{{e}_{seg}}(i)\in C$, $C=\{road,obstacle,others\}$.

 The network architecture adopts original RedNet. Considering the tradeoff of speed and accuracy, we use ResNet-34 as the feature extractor in stage two and ResNet-50 in stage one. These two networks are trained individually. After the two stages of semantic segmentation, the predicted class map is converted to a binary image by labelling all road pixels as 1 and non-road (obstacle and others) pixels as 0. It should be noted that, the semantic segmentation results will be presented by their binarized version in the following section, \ie black and white images.

\subsection{Motion Blurring for Data Augmentation}

 In actual application, as the robotic platform walks on the road or camera rotates, information coming from different sub-areas of the scene will move on the detector which will cause image blur. This phenomenon is more serious when the lighting condition is not good enough. In order to suppress segmentation errors introduced by motion blur, random motion blurring scheme is employed in this work for data augmentation. The motion blurring is commonly modeled as in~\cite{JJ07}:
\begin{equation}\label{eq:gx}
\begin{aligned}
  g(x)=(y\otimes psf)(x)
\end{aligned}
\end{equation}
 where $y\left( x \right)$ denotes the original image and $g\left( x \right)$ denotes the blurred image. $psf$ means point spread function, and $\otimes $ denotes convolution operation. For the exposure time of the proposed system is relatively short, we simplify the motion as uniform linear motion. Then $psf$ can be represented as
\begin{equation}\label{eq:psf}
\begin{aligned}
psf(x,y) = \left\{ \begin{array}{l}
\frac{1}{L},\text{if}{\;}x = y\tan \theta , \sqrt {{x^2} + {y^2}}  \le \frac{L}{2}\\
0{\;},\text{otherwise}
\end{array} \right.
\end{aligned}
\end{equation}
 where $L$ and $\theta$ denote scale and angle of motion.

\begin{figure}[tbp]%
\centering
\includegraphics[width=1\linewidth]{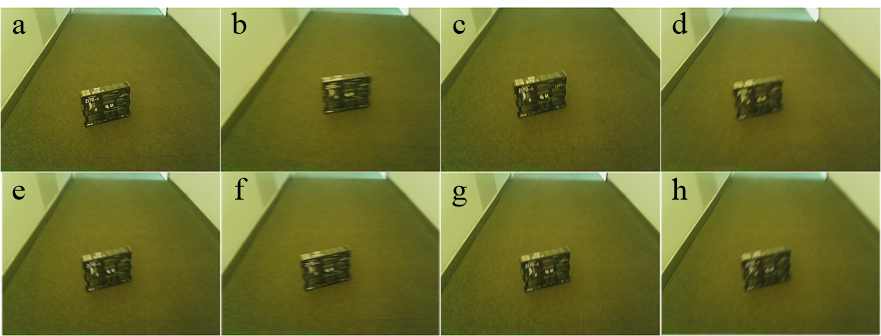}
\caption{The real clear image (a) and blurred images (b-d) and generated blur images (e-h).}
\label{fig:blur}
\end{figure}

 The motion blurring strategy is only implemented during training phase to make training data closer to actual imaging conditions. This data augmentation strategy is beneficial to obtain accurate semantic information when motion blur exists. Fig.~\ref{fig:blur} shows the comparison of real blurred images and generated images with random motion blur.

\subsection{Morphological Processing}

 We implemented morphological processing to deal with the potential imperfections in the raw binary image. Suppose that the size of binary image is $w\times h$ and all the structuring elements are square. First, the closing of the binary image by a structuring element with size ${{a}_{1}}\times {{a}_{1}}$ is performed. Then, our system performs an erosion with $a_2\times a_2$ structuring element followed by a dilation with $a_3\times a_3$ structuring element. $a_i$ is calculated by the following formula:
\begin{equation}\label{eq:ai}
\begin{aligned}
  {{a}_{i}}=f({{k}_{i}}\cdot \min (w,h))\quad i=1,2,3
\end{aligned}
\end{equation}
 where
\begin{equation}\label{eq:fx}
\begin{aligned}
  f(x)=2\cdot \left\lfloor \frac{x}{2}+1 \right\rfloor -1
\end{aligned}
\end{equation}

 The function $f(x)$ finds the closest odd integer to $x$. Assigning an odd value to ${{a}_{i}}$ makes it easier to define the origin as the center of the structuring element. The selection of ${{k}_{1}}$ depends on the largest size of obstacle we can tolerate. In our implementation, ${{k}_{1}}={1}/{80}\;$. A smaller ${{k}_{1}}$ allows the module to detect smaller obstacles but results in the increase of misdetection rate. A larger ${{k}_{1}}$ filters more misdetections and tiny obstacles out, but those not-so-small obstacles that may cause collision will also be neglected by mistake.

 The motivation for performing the following erosion and dilation is to group adjacent obstacles together so that they could be regarded as a single one. There are two reasons for this: 1) Computational complexity is reduced because the number of obstacles decreases; 2) The risk of collision caused by narrow gaps between obstacles is minimized.

 The selection of $k_2$ depends on the maximum distance of obstacles that the method should group, and the value of $k_3$ is set smaller than $k_2$ for expanding obstacles in the binary image in consideration of the robot's size. In our experiments, the combination of $k_2=1/48$ and $k_3=1/64$ shows the best performance.


\subsection{Obstacle Avoidance}

 The obstacle detection module provides the system with a smooth binary obstacle image, which is then passed to the obstacle avoidance module to determine a destination and plan a collision-free path from the current position to the destination.

\subsubsection{Local Destination Setting}

 To avoid obstacles in the present field of view, the local destination is determined firstly. The setting of a destination tries to meet the following requirements: 1) The destination must be a road pixel; 2) The road on the same horizontal line with the destination should be wide enough for the robot to pass; 3) The destination should be as far as possible within the visual range of the robot to indicate the trend of road.

 Based on these requirements, we proposed a progressive scanning method. The binary obstacle image is scanned from bottom to top. For each row, the system finds all disjoint road intervals, which are marked as their endpoints' column indices. Suppose that the module is now scanning the $i$-th row and there are $n_i$ disjoint road intervals marked as $(l_{i1},r_{i1})$, $(l_{i2},r_{i2})$, бн, $(l_{i,n_i}, r_{i,n_i})$. It's satisfied that in the $i$-th row, all pixels included in these intervals are labeled as road. Conversely, all pixels excluded in these intervals are labeled as obstacle.

 We define $d_i$ as the road breadth of the $i$-th row, which is calculated by:
\begin{equation}\label{eq:di}
\begin{aligned}
  {d_i} = \mathop {\max }\limits_{1 \le j \le n_i} \left| {{r_{ij}} - {l_{ij}}} \right|
\end{aligned}
\end{equation}

 For the second requirement, we introduce a threshold value $T = \alpha  \cdot w$, where $\alpha$ is a coefficient determining the required width of road and $w$ is the width of the binary obstacle image. In our implementation, $\alpha= 1/24$.

 With the value of $d_i$ and $T$, we can determine $g_r$, the row index of the destination. Recall that $h$ is the height of binary obstacle image and its row index increases from top to bottom, then $g_r$ is calculated as follows:
\begin{equation}\label{eq:gr}
\begin{aligned}
  {g_r} = \min \{ r|{d_i} \ge T,\forall i \in [r,h]\}
\end{aligned}
\end{equation}

 With $g_r$, the column index of the destination $g_c$ is calculated by:
\begin{equation}\label{eq:gc}
\begin{aligned}
  {g_c} = {{({l_{{g_r},m}} + {r_{{g_r},m}})} \mathord{\left/
 {\vphantom {{({l_{{g_r},m}} + {r_{{g_r},m}})} 2}} \right.
 \kern-\nulldelimiterspace} 2}
\end{aligned}
\end{equation}
 where
\begin{equation}\label{eq:m}
\begin{aligned}
  m = \mathop {\arg\max }\limits_i \left| {{r_{{g_r},i}} - {l_{{g_r},i}}} \right|
\end{aligned}
\end{equation}

 Finally, the pixel at row $g_r$ and column $g_c$ is determined as the destination, which will be used for path planning in the next section. Fig.~\ref{fig:pplan} gives some examples on destination settings. Furthermore, our destination setting algorithm is suitable for parallel computing thus implemented on GPU for real-time application.

\subsubsection{Local Path Planning}

 With the local destination determined in the previous section, the obstacle avoidance module is now prepared to plan a path using Artificial Potential Field (APF) method~\cite{APF}.

 APF, first used by Khatib, is an algorithm that can best realize our target of planning collision-free route for robots in real-time. It can construct an artificial potential field, where obstacles have repulsive potential fields repelling the robot, while the designated destination has attractive potential field pulling the robot. Consequently, our robot is under the resultant force and steered towards the destination.

 Suppose that there are $n$ obstacles in the binary image named $o_1$, $o_2$,..., $o_n$. The distance and angle between $o_i$ and the robot are marked as $d_i$ and $\theta_i$. The distance and angle between the destination and the robot are marked as $d_g$ and $\theta_g$. And we introduce $\mu_r$ and $\mu_a$ as the repulsive and attractive scaling factors respectively. Then the repulsive force vector $F_r$ and attractive force vector $F_a$ on the robot is calculated as follows:
\begin{equation}\label{eq:fr}
\begin{aligned}
  {F_r} = {\mu _r}\sum\nolimits_i {\frac{1}{{d_i^2}}}  \cdot \left( {\sin {\theta _i},\cos {\theta _i}} \right)
\end{aligned}
\end{equation}
\begin{equation}\label{eq:fa}
\begin{aligned}
  {F_a} = {\mu _a}\frac{1}{{d_g^2}} \cdot \left( {\sin {\theta _g},\cos {\theta _g}} \right)
\end{aligned}
\end{equation}
 Therefore, the resultant force $F$ on the robot equals to:
\begin{equation}\label{eq:fa}
\begin{aligned}
  F = {F_a} - {F_r}
\end{aligned}
\end{equation}

\begin{figure}[tbp]%
\centering
\includegraphics[width=1\linewidth]{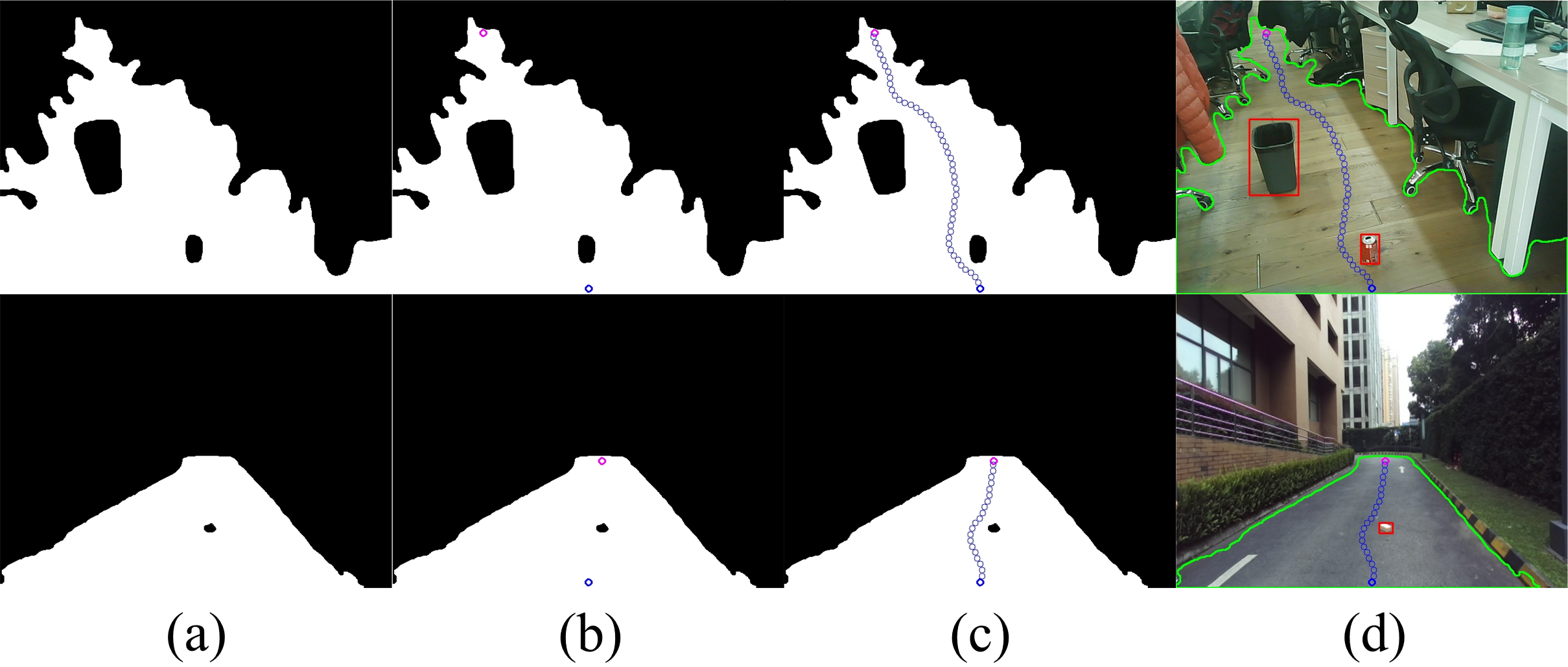}
\caption{Path planning based on binary image. (a) Morphological processed image. (b) Destination setting (colored pink). (c) Path planning. (d) Mapping to original image.}
\label{fig:pplan}
\end{figure}

 In each step, the algorithm calculates the resultant force $F$ that affects the robot's forward direction, then steers the robot to the next position. This procedure will be repeated until the robot reaches its destination. If the path planning module cannot find a collision-free path to the destination, the robot will rotate itself by 15 degrees and reset a destination then try to plan a path to the new destination. Some results of path planning procedure are shown in Fig.~\ref{fig:pplan}.

\section{Evaluation}

 In this section, we evaluate the proposed method both on SUN RGB-D dataset and Cityscapes dataset for indoor and outdoor scenarios respectively. Besides, we also evaluation the performance of obstacle avoidance on a new established small obstacle dataset.

 SUN RGB-D~\cite{SUN} dataset consists of 10335 indoor RGB-D images with pixel-wise semantic annotations of 37 object classes. It has a default trainval-test split which is composed of 5285 images for training and validation and 5050 images for testing.

 Cityscapes~\cite{CITY} is a large-scale dataset for urban scene semantic understanding. It contains footages of street scenes collected from 50 different cities, at a frame rate of 17 fps. The train, validation, and test sets contain 2975, 500, and 1525 footages, respectively. Each footage has 30 frames, where the 20th frame is annotated by 30 semantic classes with pixel-level ground-truth labels.

 The new small obstacle dataset is captured in indoor and outdoor scenarios respectively with Orbbec Astra and ZED with a height of about 1.2m. The spatial resolutions of RGB-D are both $640\times480$. The captured data are pixel-wise labeled as road, obstacle, and others. Besides, for each image, walking routes are labeled by five people. The reasonable ground truth of path plan is obtained by their mean. Samples for indoor and outdoor scenarios are 2200 and 2000. The obstacles we choose are arbitrary objects with random color or shape and with the size ranging from $5cm\times5cm\times5cm$ to $50cm\times50cm\times50cm$ that may hinder walking of robots, such as trash can, carton, brick, and other small objects easily falls on the road.

\subsection{Training Scheme}

 The key to obstacle avoidance based on semantic segmentation is precise segmentation result of road area. For the two public dataset, there is no concept of obstacle, so their original classes are mapped to a new list in both training and inference of stage one. For the SUN RGB-D dataset, the principles of mapping are similar to the structure class of NYUDv2 dataset~\cite{NYUD}, floor, wall, window, and ceiling are reserved as scene classes. A few classes that are adjacent to the scene classes and have similar depth with the former are merged, such as floor-mat and floor, or blinds and window. Other classes are divided to furniture and objects. Furniture are large objects that cannot be easily moved, objects are small objects that can be easily carried. For the Cityscapes dataset, the mapping principles are implemented as the original large categories, \ie flat, construction, object, and so on. The new small obstacle dataset is only used during inference.

 According to our statistics, most of blur kernel sizes are within scope of 5 pixels. In training, each sample is added by random linear blurring before fed to the model, with a kernel size of 3, 5, or 7 pixels and a blur direction from 0 to $\pm$180 degree. The original larger images are cropped to $640\times480$ randomly for the proposed network both in training and inference. ResNet-50 and ResNet-34 pre-trained on ImageNet object classification dataset are used as the encoder for the two stages. Training for the two stages are performed individually. Images of SUN RGB-D dataset are translated to calculate optical flow with original images in training.

 The proposed two-stage semantic segmentation architecture is implemented on the Pytorch deep learning framework. Stochastic gradient descent (SGD) is applied for both of the networks. The initial learning rate is set to 0.002 for stage one, and 0.0002 for stage two. Both the learning rates are decayed by a factor of 0.8 in every 50 epochs. The networks are trained on 4 Nvidia Tesla P100 GPUs with a batch size of 10 until the losses do not further decrease.

\subsection{Evaluation Methodology}

\begin{table}[tbp]
\center
  \caption{Semantic segmentation performance of indoor scenario}
  \label{tb:indoor}
\begin{center}
\begin{tabular}{c|c|c|c|c}
\hline
Model & mIoU$_1$ & mIoU$_2$ & ODR & NOFP\\
\hline
SegNet & 58.1 & -- & 68.4 & 9.6\\
FuseNet & 70.2 & -- & 75.7 & 7.7\\
RedNet & 74.5 & -- & 84.1 & 4.5\\
Ours & 75.7 & 98.2 & 95.2 & 2.7\\
Ours(+blur) & 76.2 & 98.5 & 95.8 & 2.3\\
Ours(+blur+flow) & \textbf{76.9} & \textbf{99.2} & \textbf{96.3} & \textbf{2.2}\\
\hline
\end{tabular}
\end{center}
\end{table}

\begin{table}[tbp]
\center
  \caption{Semantic segmentation performance of outdoor scenario}
  \label{tb:outdoor}
\begin{center}
\begin{tabular}{c|c|c|c|c}
\hline
Model & mIoU$_1$ & mIoU$_2$ & ODR & NOFP\\
\hline
PSPNet & 87.8 & -- & 71.4 & 8.2\\
DeepLabv3+ & 89.3 & -- & 77.9 & 6.7\\
Ours & 91.1 & 96.4 & 92.7 & 5.0\\
Ours(+blur) & 91.5 & 96.6 & 92.9 & 4.7\\
Ours(+blur+flow) & \textbf{92.1} & \textbf{97.2} & \textbf{93.8} & \textbf{4.2}\\
\hline
\end{tabular}
\end{center}
\end{table}

 The evaluation consists of two parts: obstacle segmentation and obstacle avoidance. The obstacle segmentation is performed on the default testing set of SUN RGB-D dataset (indoor scenario) and validation set of Cityscapes dataset (outdoor scenario) for the mapped categories of stage one, and on the small obstacle dataset for road/obstacle. The obstacle avoidance is performed on the small obstacle dataset. We evaluate semantic segmentation accuracy on pixel-level, and instance-level for obstacles of stage two. We adopt mean intersection-over-union (mIoU) scorce as the criteria of pixel level. mIoU$_1$ and mIoU$_2$ denote the performance of stage one and stage two respectively. The instance-level accuracy is measured by obstacle detection rate (ODR) and non-obstacle false positives (NOFP). If more than 50\% pixels of a predicted obstacle instance are overlapped with the ground truth, it's a success prediction. If there's no pixel of a predicted instance is overlapped with obstacle ground truth, it's a false prediction. ODR is defined as:
\begin{equation}\label{eq:odr}
\begin{aligned}
  \text{ODR}=\text{SPI}_{obs}/\text{TI}_{obs}
\end{aligned}
\end{equation}
where $\text{SPI}_{obs}$ is the quantity of success predictions, $\text{TI}_{obs}$ the total obstacle instances. NOFP if defined as:
\begin{equation}\label{eq:nofp}
\begin{aligned}
  \text{NOFP}=\text{FPI}_{obs}/\text{TF}
\end{aligned}
\end{equation}
where $\text{FPI}_{obs}$ is the quantity of false prediction instances, and TF the total test frames. The obstacle avoidance quality is measured by Hausdorff distance of planned path with the ground truth path.

\subsection{Evaluation of Obstacle Segmentation}

\begin{figure}[tbp]%
\centering
\includegraphics[width=1\linewidth]{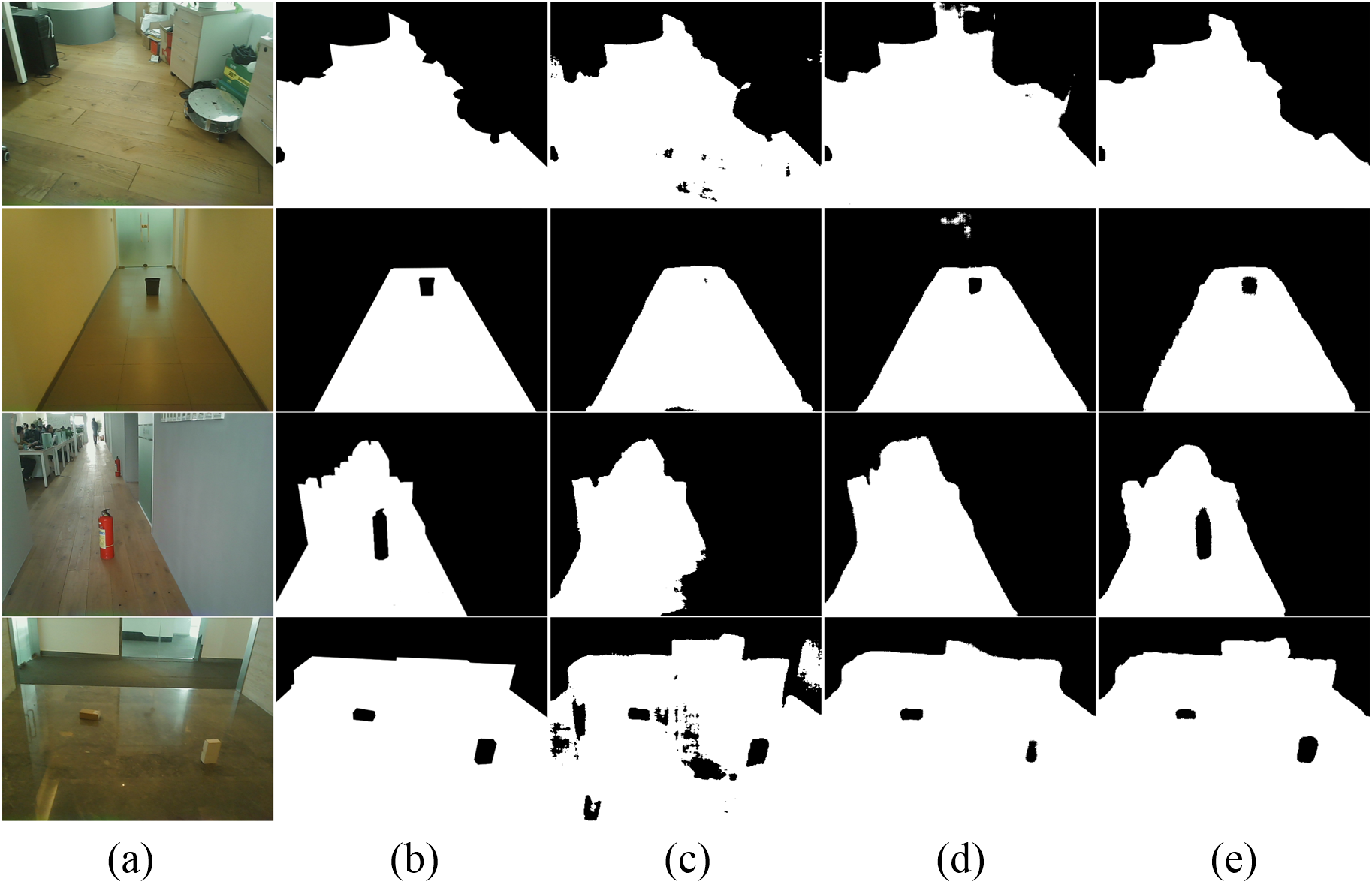}
\caption{Obstacle segmentation results of different models. (a) The input image. (b) Ground truth of semantic segmentation. (c) Result of Segnet. (d) Result of RedNet. (e) Result of ours.}
\label{fig:indoor}
\end{figure}

\begin{figure}[tbp]%
\centering
\includegraphics[width=1\linewidth]{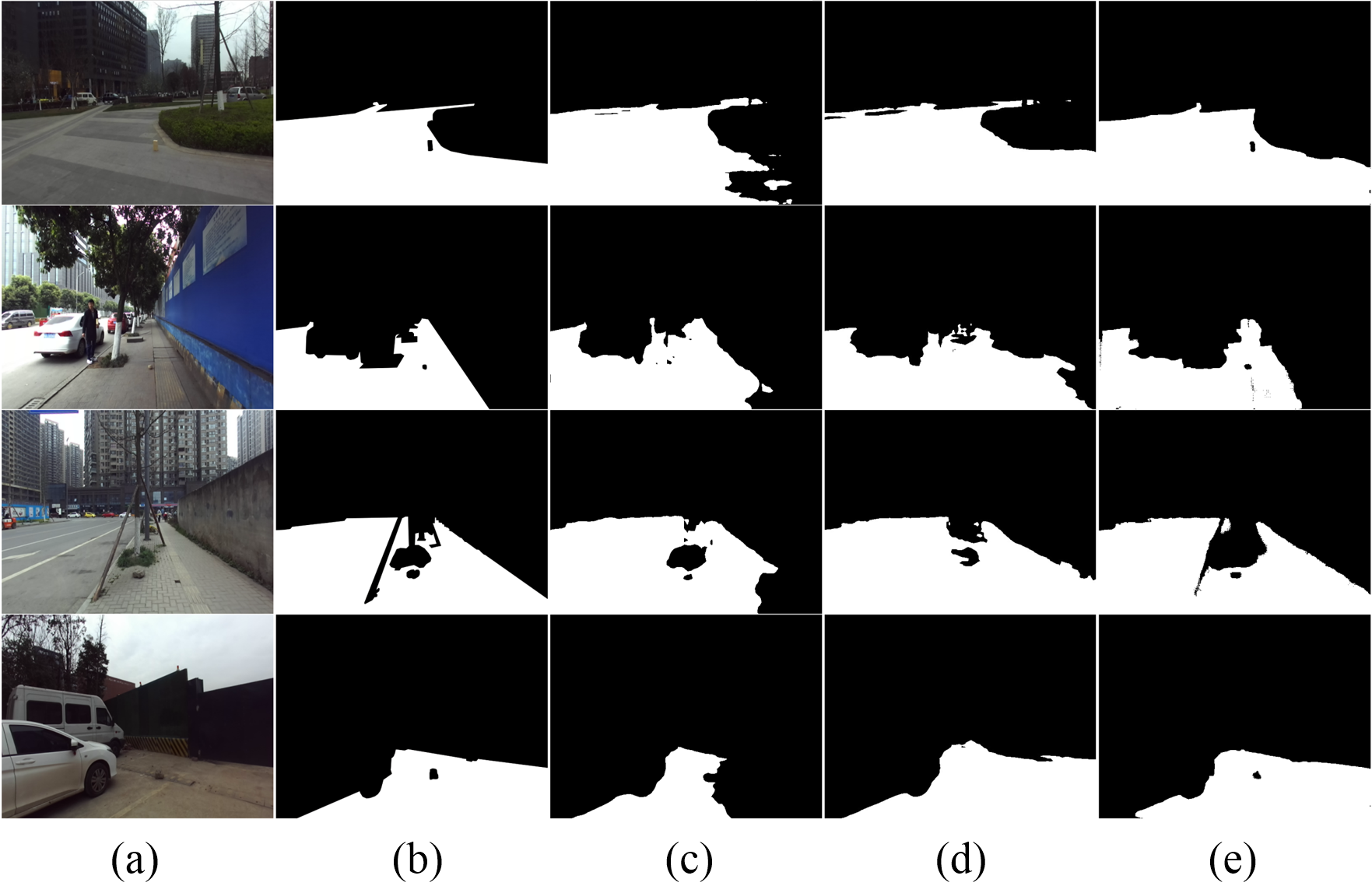}
\caption{Outdoor obstacle segmentation results of different models. (a) The input image. (b) Ground truth of semantic segmentation. (c) Result of PSPNet. (d) Result of DeepLabv3+. (e) Result of ours.}
\label{fig:outdoor}
\end{figure}

 Table~\ref{tb:indoor} and~\ref{tb:outdoor} shows the performance comparison of our indoor and outdoor models with some state-of-art semantic segmentation based on RGB and RGB-D data. The proposed method outperforms other semantic segmentation architectures both in indoor and outdoor scenarios. After the augmentation of random motion blurring, the mIoU elevates 1.2\% and 0.4\% in two scenarios respectively. Furthermore, accuracy of instance-level is also improved. Due to the introduce of optical flow supervision, the performance of obstacle segmentation is better and stable.

\begin{figure}[tbp]%
\centering
\includegraphics[width=1\linewidth]{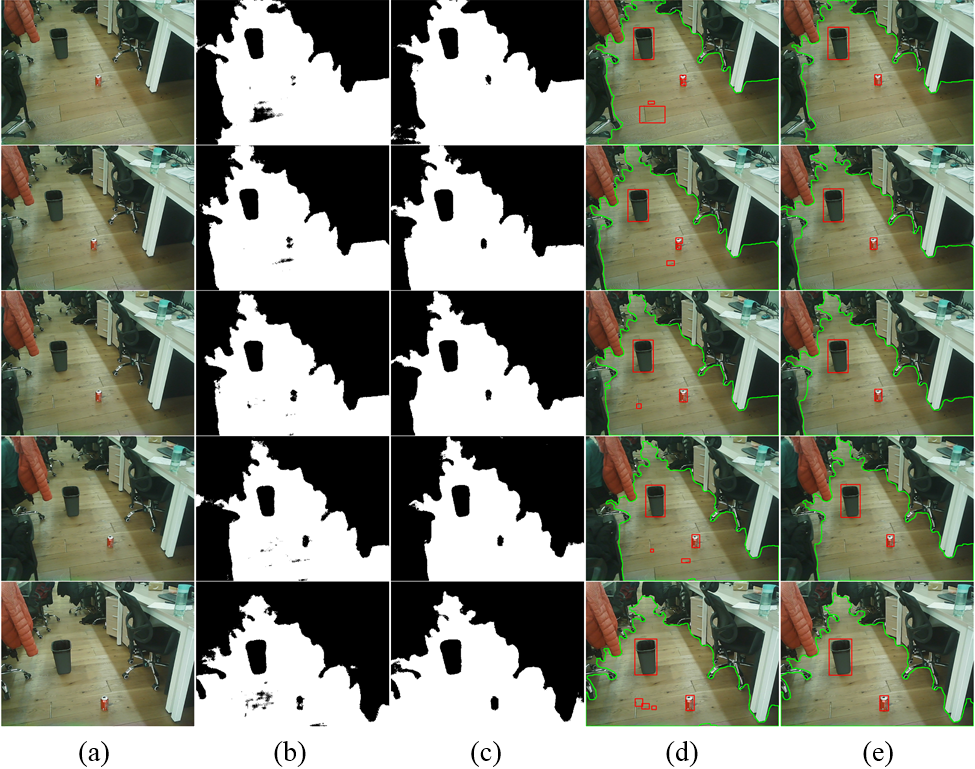}
\caption{Segmentation performance of our method on consecutive frames. (a) The input image. (b) and (c) are segmentation result without and with temporal consistency supervision. (d) and (e) are the obstacle detection result of (b) and (c).}
\label{fig:cons}
\end{figure}

 Some obstacle segmentation results are shown in Fig.~\ref{fig:indoor} and~\ref{fig:outdoor}. As can be seen, obstacles and road areas are both segmented more accurate compared with other architectures, which is crucial to the following obstacle detection and avoidance. Moreover, the proposed method can detect very small obstacles successfully, while other methods fail to. Fig.~\ref{fig:cons} presents the segmentation performance of our method on consecutive frames with and without optical flow supervision. As can be seen, with the supervision optical flow, the segmentation results are relatively stable, in other words, the temporal consistency is kept.

\begin{table}[tbp]
\center
  \caption{Path planning accuracy}
  \label{tb:path}
\begin{center}
\begin{tabular}{c|c|c}
\hline
Model & Indoor & Outdoor\\
\hline
Stereo Method~\cite{GB17} & $\sim$0.6m & --\\
Ours & 0.15m & 0.27m\\
\hline
\end{tabular}
\end{center}
\end{table}

\subsection{Evaluation of Obstacle Avoidance}

\begin{figure}[tbp]%
\centering
\includegraphics[width=1\linewidth]{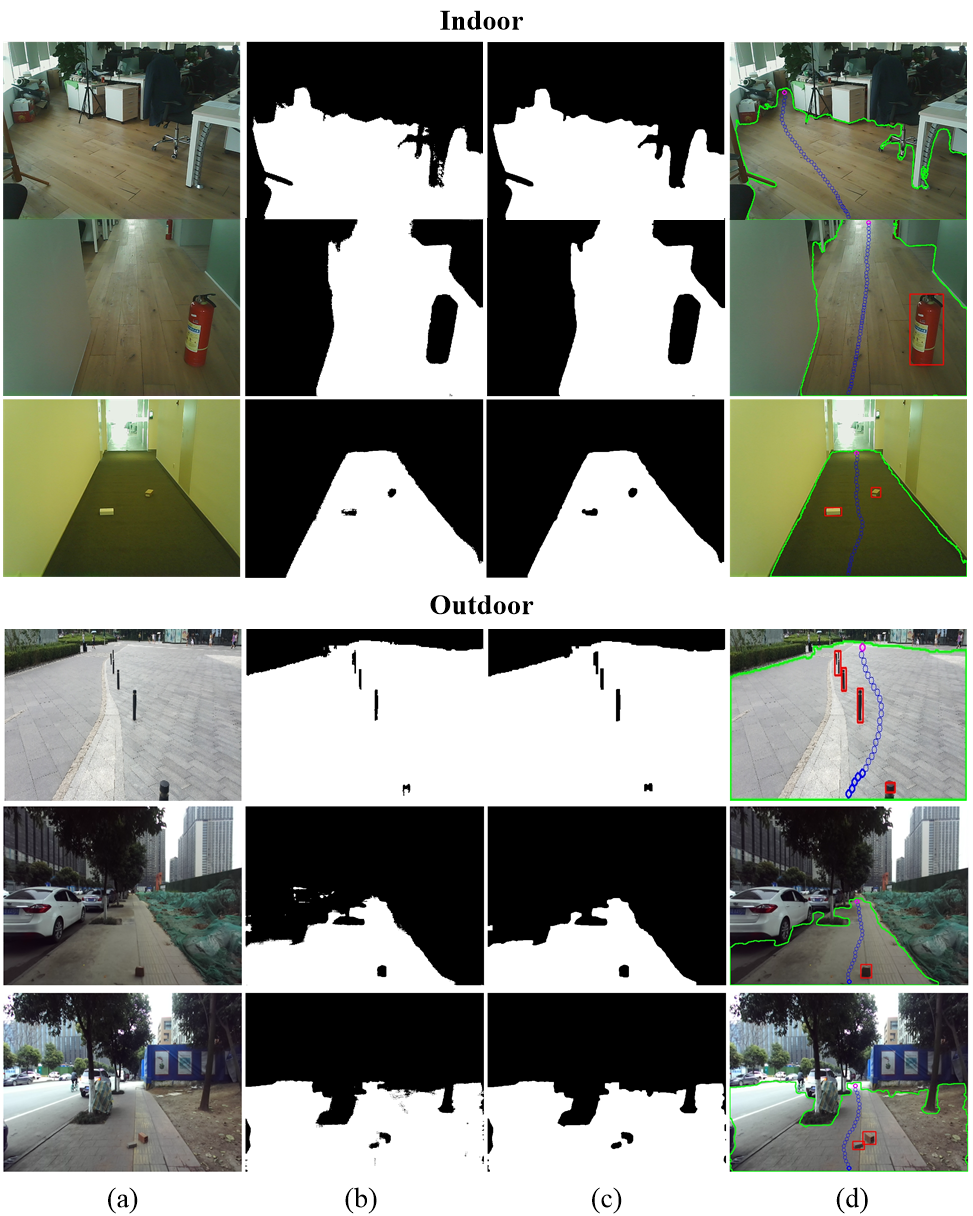}
\caption{Obstacle avoidance results for indoor and outdoor scenarios. (a) image (b) segmentation (c) morphological processing (d) path planning}
\label{fig:res}
\end{figure}

 The obstacle segmentation results are morphological processed to make path planning. Table~\ref{tb:path} presents the path planning performance evaluated by Hausdorff distance. The paths made by the proposed method is very close to the ground truth and surpass stereo based method. Fig.~\ref{fig:res} shows examples of our obstacle avoidance results. The first column shows the original input color images, the second the binary images calculated by the deep segmentation network, the third the binary images after morphological processing, and the last the result of our system with road area contoured by green line, obstacles marked by red bounding, destination marked by pink circle and path marked by several blue circles. Five thick blue circles, closest to the bottom of the images, indicate five steps calculated in one calculation, while the rest are for demonstration. Above evaluation results indicate the proposed method can effectively detect obstacles and make reasonable path planning for robots in various complicated scenarios.

\section{Conclusion}

 In this study, we proposed a new architecture to automatically create walking routes with RGB-D images for road robots based on deep semantic segmentation neural networks. Two-stage segmentation with motion blurring augmentation and optical flow supervision is presented to acquire more accurate and stable obstacle segmentation, morphological processing is applied to refine the obstacle segmentation, and based on the accurate obstacle map, the local path planning is done to produce the collision-free path. Experimental results show that the proposed method works well under various scenarios whether indoor or outdoor, even with small obstacles or capture blurs. It is the two-stage segmentation that improves small obstacle detection accuracy, and the blurring based data augmentation and optical flow supervision that improves the stability.

{\small
\bibliographystyle{ieee}
\bibliography{egbib}
}

\end{document}